\documentclass[pdflatex,referee,sn-mathphys]{sn-jnl}

\jyear{2021}%

\theoremstyle{thmstyleone}%
\theoremstyle{thmstyletwo}%

\theoremstyle{thmstylethree}%

\begin{document}

\title[A Unified Framework for Attention-Based Few-Shot Object Detection]{A Unified Framework for Attention-Based Few-Shot Object Detection}


\author[1,2]{\fnm{Pierre} \sur{Le Jeune}}\email{pierre.le-jeune@cose.fr}

\author[1]{\fnm{Anissa} \sur{Mokraoui}}\email{anissa.mokraoui@univ-paris13.fr}

\affil[1]{\orgdiv{L2TI}, \orgname{Université Sorbonne Paris Nord}}

\affil[2]{\orgname{COSE}}


\abstract{Few-Shot Object Detection (FSOD) is a rapidly growing field in
computer vision. It consists in finding all occurrences of a given set of
classes with only a few annotated examples for each class. Numerous methods have
been proposed to address this challenge and most of them are based on attention
mechanisms. However, the great variety of classic object detection frameworks
and training strategies makes performance comparison between methods difficult.
In particular, for attention-based FSOD methods, it is laborious to compare the
impact of the different attention mechanisms on performance. This paper aims at
filling this shortcoming. To do so, a flexible framework is proposed to allow
the implementation of most of the attention techniques available in the
literature. To properly introduce such a framework, a detailed review of the
existing FSOD methods is firstly provided. Some different attention mechanisms
are then reimplemented within the framework and compared with all other
parameters fixed.}


\maketitle

\section{Introduction}
\label{sec:intro}
Few-Shot Object Detection (FSOD) is a challenging problem that aims to find all
occurrences of a class in an image given only a few examples. Impressive
progress has been made in object detection in the past decade, mostly because of
deep convolutional networks (see e.g. \cite{ren2015faster,redmon2016you}).
Current state-of-the-art performs high-quality detection, but it requires large
annotated datasets and days of training to achieve that quality. Often, such
requirements could not be met, and it is quite hard to achieve good performance
for a specific task. The few-shot learning trend focuses specifically on this
kind of use cases where data is scarce. Most few-shot approaches rely on a
two-steps training strategy. First, the model is trained on a large dataset,
slightly different from the desired task, to acquire general knowledge. Second,
a fine-tuning step refines the model to learn the actual task. Of course, plenty
of techniques have been introduced so that the fine-tuning does not disrupt base
learning (see e.g. \cite{kirkpatrick2017overcoming}). This has been extensively
studied for classification in the past years. However, detection is a more
challenging task than classification and has been only tackled recently from a
few-shot perspective.

Current FSOD state-of-the-art is mainly based on attention mechanisms, which aim
at extracting information about the task (i.e. semantic information about the
classes that should be detected) from support examples. This information allows
the network to condition the detection on the examples. It allows the network to
adapt on the fly, provided with a few annotated images and a short fine-tuning
step. A seminal work in this direction is presented in reference
\cite{kang2019few} which reweights features from the query image (i.e. the
images in which the model performs the detection) with the features extracted
from the support images. Plenty of methods based on the same idea have been
introduced (see e.g. \cite{deng2020few,fan2020few,wallach2019one,li2020one}).
Although the attention mechanisms, deployed in most FSOD methods, might differ
from the original one proposed in \cite{kang2019few}, the main principle remains
the same. Indeed, it combines information from the query image and the support
set to detect only the objects annotated in the support.

The FSOD field is rapidly growing, and most new papers propose a novel attention
technique. However, there are a lot of design choices that can be considered to
address the FSOD problem. First, the detection framework (e.g. Faster R-CNN
\cite{ren2015faster} or YOLO \cite{redmon2016you}) and its backbone (e.g.
ResNet-50 or 101), then the different loss functions (e.g. L1, IoU, Focal Loss)
to train each part of the network and finally all the hyperparameters that are
tied to these methods (i.e. from the learning rate to the class splits for
evaluation). All this makes the comparison between FSOD methods difficult. In
this paper, the attention mechanism is considered as the heart of FSOD because
it combines general features extracted from the input image and the conditioning
features extracted from the support examples. Of course, it should be noted that
not all works on FSOD are attention-based. Indeed, there exist some papers that
use metric learning to solve FSOD (see e.g.
\cite{karlinsky2019repmet,jeune2021experience,sun2021fsce}). Instead of
combining query and support features, a generic embedding function is learned.
The detection is then obtained from the comparison of the query and support
embeddings. There are also approaches based only on fine-tuning. Although quite
straightforward these methods generally do not perform as well as
attention-based ones.

The main goal of this paper is to review a part of the attention-based FSOD
methods and compare them fairly. To do so, a modular attention framework is
proposed. It allows regrouping most attention-based methods under the same
notations. Specifically, this framework is divided into three parts: spatial
alignment, global attention, and fusion layer. This separation helps to easily
implement the different mechanisms and facilitates the comparison. Most
importantly, this makes it possible to completely fix the other parameters and
design choices without reimplementing each method. That way, a fair comparison
of the performance of the different mechanisms can be obtained. The modular
splitting also allows a deep understanding of what component of the attention is
key for FSOD and gives insights to build better approaches. Moreover, to help
the development of new techniques for FSOD and future comparisons, the code of
the proposed framework will be made
available\footnote{\url{https://github.com/pierlj/aaf_framework}}. To our
knowledge, there exist two papers that review FSOD methods
\cite{huang2021survey, jiaxu2021comparative}. However, they focus on general
FSOD while this work targets attention-based FSOD.

The rest of this paper is organized as follows. First, a review of existing
methods for FSOD is conducted in Section \ref{sec:related}. A formal description
of the attention framework is then proposed in Section \ref{sec:framework}.
Section \ref{sec:comparison} confirms the flexibility of the framework with a
comparative analysis of some existing methods for FSOD. All these methods have
been reimplemented through the proposed framework, achieving similar results to
those reported in the original papers.  Some insights about the key elements of
the attention mechanisms are then provided. Section \ref{sec:conclusion}
concludes this paper.


\section{Review of Existing Work on FSOD}
\label{sec:related}
This section reviews existing work on few-shot object detection. It begins with
a summary of general object detection methods. Next, it presents the
main principles of few-shot learning, and finally, it explains how these
principles are applied to object detection in the scientific literature.

\subsection{Overview of Object Detection Methods}
Object detection, among other computer vision challenges, has made impressive
progress with the rise of Convolutional Neural Networks (CNN). Some previous
works, such as \cite{girshick2014rich, girshick2015fast}, leverage CNNs to
classify Region of Interest (RoI) generated by classical methods (e.g. selective
search). However, these methods are slow and achieve relatively low performance.
YOLO \cite{redmon2016you} and Faster R-CNN \cite{ren2015faster} have been the
first fully convolutional approaches for object detection and are still the root
of most recent work in this field. These two architectures are the most
representative examples of the two kinds of detectors that exist: one-stage
detectors with a trade-off on speed and two-stages detectors with a trade-off on
accuracy. Plenty of improvements have been introduced over these methods. On the
one hand, for one-stage detectors, Focal Loss \cite{lin2017focal} improves
training balance between background and foreground. CenterNet
\cite{duan2019centernet} and CornerNet \cite{law2018cornernet} propose new ways
to predict boxes coordinates. FCOS \cite{tian2019fcos} gets away from the
concept of anchors boxes (a predefined set of boxes introduced both in YOLO and
Faster R-CNN). On the other hand, two-stages detectors have been improved with
pyramidal features extraction by \cite{lin2017feature}, improving detection of
objects of various sizes. Mask R-CNN \cite{he2017mask} improves performance
by adding an instance segmentation branch to Faster R-CNN. Recently, various
works exploit self-attention mechanisms to increase detection quality.
Dynamic Head combines three different types of attention, based on scale,
location and current task \cite{dai2021dynamic}. Pushing even further, DETR
\cite{carion2020end} replaces the convolutional regressor and classifier with
transformers.

\subsection{Basic Concepts on Few-Shot Learning}
\label{sub:basic}
Deep learning based methods achieve impressive results when provided with
sufficient training data. However, it is often quite challenging to gather such
a dataset for a given use case. Hence, Few-Shot Learning (FSL) aims at learning
tasks from only limited data. The main principle is to learn generic knowledge
from large-scale training on a task similar to the actual problem. The model
is then adapted to perform the final task from the scarce data available.

In FSL literature, a task is defined as $K$-shots, $N$-ways learning when the
training set only contains $K$ examples for each of its $N$ classes. It is
common to introduce the support and query sets for a given task. The support
contains the available examples: $K$ images for each of the $N$ classes. It
allows the network to adapt to the task by extracting relevant information from
the examples. The query set contains images from the same classes and is used
either for inference or training. For clarity, query and support images will
refer to elements of the query and support sets respectively. Moreover, to
assess the performance of FSL the set of classes is divided in base (or seen)
and novel (or unseen) classes. Train and test classes can also be found in the
literature. A large quantity of examples is available for base classes that
serve during base training while for novel classes, only $K$ examples are
available per class. Novel classes are presented to the network during
fine-tuning and evaluation only. This allows assessing the generalization
capabilities of the methods.

\begin{figure}[h]
    \centering
    \includegraphics[width=0.5\textwidth]{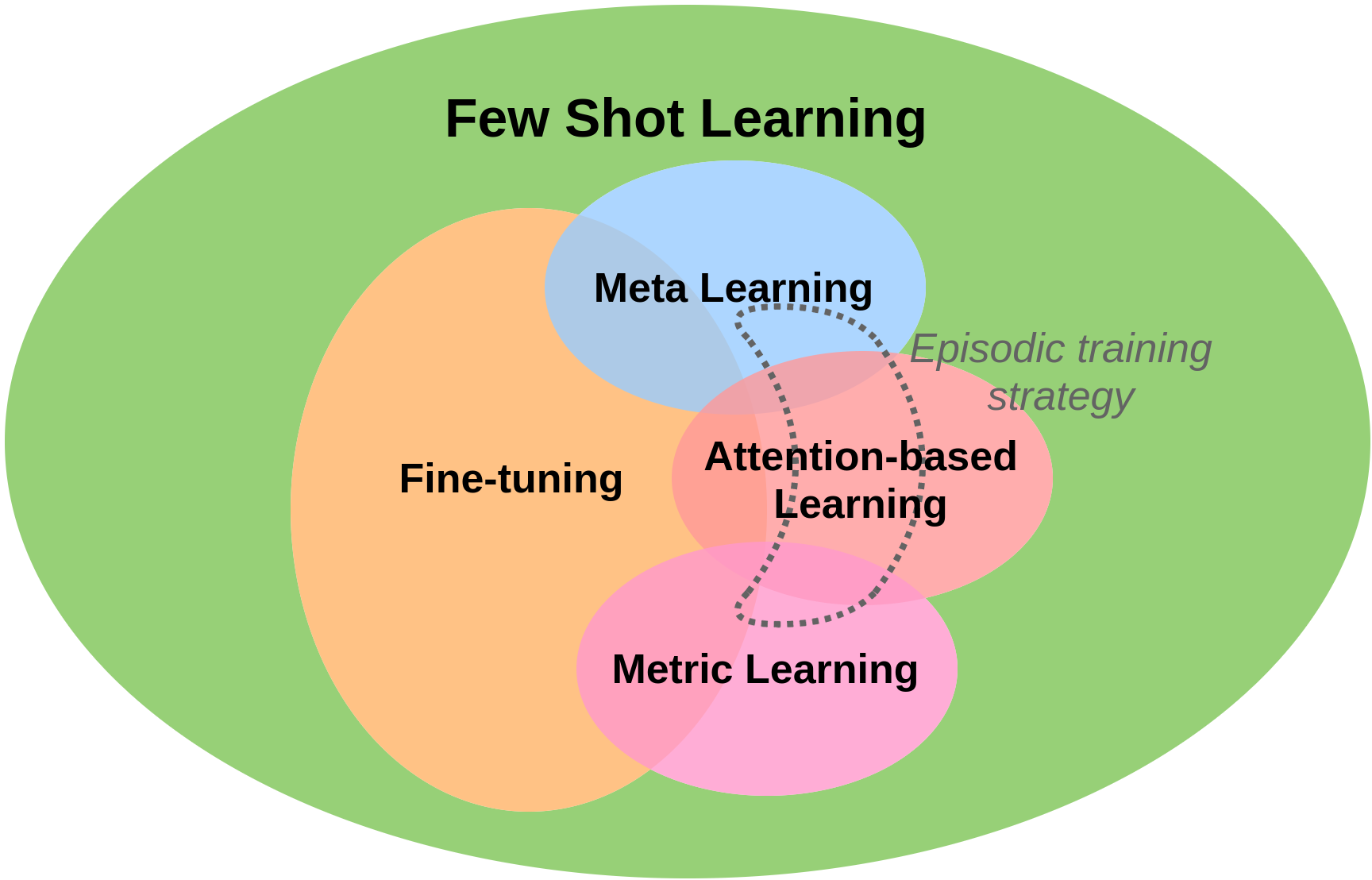}
    \caption{Few-shot learning methods can be divided into 4 main categories:
    fine-tuning, meta learning, attention-based learning and metric learning.
    Those are not completely separated, in particular, fine-tuning is part of
    the training strategy of many FSL methods. Episodic task training is also
    a widespread strategy for FSL.}
    \label{fig:fsl_classification}
\end{figure}

As summarized in Figure \ref{fig:fsl_classification},  FSL approaches can be
classified as follows: fine-tuning, meta-learning, metric learning and
attention-based. This classification could be refined as in
\cite{wang2020generalizing} which offers a complete review of FSL.

\noindent
\textbf{Fine-tuning \textendash} This is the simplest way to tackle FSL. It
consists in training the model on a large dataset with base classes examples
only and then fine-tune it with a limited number of examples of the novel
classes. While conceptually simple, these approaches prove to be effective.
Nonetheless, they are often prone to catastrophic forgetting
\cite{kirkpatrick2017overcoming}: performance drops on base classes after
fine-tuning. Plenty of tricks have been introduced to alleviate this issue.
Fine-tuning on its own is not very powerful for FSL, but it is part of most
other methods as their training strategy (see Figure
\ref{fig:fsl_classification}).

\noindent
\textbf{Meta-learning \textendash} It attempts to learn models that can quickly
adapt to a task. This is often performed by training two separate models: a
meta-learner and a student (or learner). The meta-learner's goal is to help
the learner to train on new tasks. This can be achieved in different ways. For
instance, in reference \cite{ravi2016optimization}, a meta-learner that directly
outputs weights updates for the student is proposed. Similarly, reference
\cite{finn2017model} proposes to only output initial weights for the learner. 
These techniques rely on a two-level optimization procedure. At the lower level,
a task is selected (e.g. for classification: choosing a random subset of
classes), and then the learner is trained for this task. At the higher level,
the performance of the student is assessed on the current task and serves as a
cost function for the meta-learner. This episodic training strategy improves
adaptability to new tasks. However, these methods do not scale very well as the
meta-learner often needs to be quite larger than the actual task learner.

\noindent
\textbf{Metric Learning \textendash} The goal of metric learning is to learn a
generic embedding function from the base dataset, such that the embedding space
is structured semantically, making it easier to distinguish between
classes. To do so, a metric function is learned over the data manifold (or
equivalently an embedding function that maps the data points into a Euclidean
space). Input images are then classified according to the distance between their
representations and the representations of the support examples in the embedding
space. This has been introduced in the context of few-shot learning by
Prototypical Networks \cite{snell2017prototypical}. 
Similarly, two separate networks can be trained to embed the support and query
sets separately (e.g. \cite{vinyals2016matching}). Following these two methods,
a series of improvements have been developed. For instance, in
\cite{sung2018learning} a second network is trained to compute the similarity
between prototypes and query embedding. 

\noindent
\textbf{Attention-based Methods \textendash} Finally, there are methods similar
to metric learning but which do not directly compare the query and support
representations. Instead, the support representation is used to change the
parameters of the model on the fly to adapt it to new tasks (i.e. new
classes). The original idea has been proposed by \cite{bertinetto2016learning}.
A classification network is trained at the same time as a \textit{learnet} whose
purpose is to output some weights for the main network from the support images.
That way, the network has dynamic weights and can adapt to new classes.
Practically, the \textit{learnet} is trained to output kernel weights adapted to
the task from the support examples. This kernel is then plugged into the main
detection network in a dynamic convolutional layer. 
This can be seen as an attention mechanism between a query image and support
images, hence the Attention-based methods. Note that this is referred to as
\textit{modulation} in \cite{huang2021survey}. The query features are reweighted
by the support features, with a channel-wise multiplication.

Usually, attention highlights features that are relevant to the task. In the
case of self-attention (i.e. attention on the features themselves), this can be
achieved by multiplying channel-wise a globally pooled feature vector with the
map itself, as proposed by \cite{hu2018squeeze}. It can also be with spatially
distant features of the same image as in non-local neural networks
\cite{wang2018non} or Visual Transformers (ViT) \cite{dosovitskiy2020image}. But
attention can also be computed with features coming from different images. In
this case, features from external images are highlighted in the original map,
this can be referred to as external or cross attention. This is particularly used
for FSL to adapt query features to support examples. For instance, in
Cross-transformers \cite{doersch2020crosstransformers}, the authors leverage
transformer-like attention to spatially align the supports to the query. This
makes the model more robust to intra-class variance and support choices.

\subsection{Review of Few-Shot Object Detection Methods}
Few-Shot Learning has been extensively studied in the light of classification
tasks. For object detection however, the scientific literature is scarcer. This
section aims to present some works proposed for FSOD and to compare them. They
are divided into three different groups: transfer learning, metric learning, and
attention-based learning. To our knowledge, there is no work completely based on
meta-learning that tackles FSOD.

\subsubsection{Fine-tuning}
Low Shot Transfer Detector (LSTD) is a pioneer work on FSOD \cite{chen2018lstd}.
It proposes to first train a detector (Faster R-CNN) on a base dataset and then
fine-tune it on a novel set containing only some examples of the novel classes.
To prevent catastrophic forgetting, the authors introduced two regularization
losses so that the network produces similar outputs for the base classes during
fine-tuning. Closely related, reference \cite{wang2020frustratingly} leverages
the same idea without any additional loss. Instead, they choose to freeze all
the network weights after base training except for the last classification and
regressions layers. Reference \cite{wu2020multi} also proposes a basic
fine-tuning strategy, but instead of freezing the network, they leverage a
multiscale refinement branch, used only for training, to better guide the
classification branch of the network. This way, a better balance between
positive and negative samples is achieved, making base training and fine-tuning
more efficient. Another method is proposed in \cite{fan2021generalized} where
two Faster R-CNN are trained: one on base classes and one on all classes (base
and novel), as a fine-tuned version of the first one. Their main contribution is
that both detectors are used together to output predictions. First, the
objectness maps of base and novel detectors are combined so that the Region
Proposal Network (RPN) outputs proposals both for base and novel classes. Next,
the predictions of the base and novel heads are combined to get the bounding
boxes for all classes. The authors also introduce a consistency loss  so that
the novel detector is encouraged to retain knowledge about base classes and
alleviate catastrophic forgetting.

\subsubsection{Metric Learning}
For few-shot classification, metric-learning based methods are probably the most
widespread. However, for detection, only a few works are based on this
technique. RepMet \cite{karlinsky2019repmet} is one of the first, it consists in
learning class representative vectors while training a modified Faster R-CNN
detector. Closely related, \cite{zhang2021pnpdet} learns prototypes vectors as
well as scale factors inside the CenterNet framework. These vectors are used in
the classification head of the detector as class prototypes. The difference is
that the vectors are learned through training and not simply computed from
examples like in Prototypical Faster R-CNN \cite{jeune2021experience}. The
authors of this method proposed to embed a vanilla prototypical network into
Faster R-CNN not only in the detection head but in the RPN as well. This
prevents the RPN to specialize in base classes and output better proposals.
Nevertheless, authors relate mitigated results with this method. They
hypothesize that computing prototypes from examples may not be the best
detection strategy as the background is often dominant in the image, which can
lead to biased prototypes and shortcut learning.

Reference \cite{sun2021fsce} is halfway between RepMet and Prototypical Faster
R-CNN. It computes the prototypes directly from the example but only in the
second stage of the network. In addition, it leverages a contrastive supervised
loss function to learn a semantically aware embedding space. These methods are
all similar in how the class prototypes influence the classification of RoI. On
the contrary, reference \cite{wu2021universal} uses the prototypes as
reweighting vectors to enhance class-specific features of the embedded RoI. This
is astride metric learning and attention-based methods, but it is presented from
a metric learning perspective by its authors.

\subsubsection{Attention-based}
\label{sec:attention_fsod}
To address the shortcoming of metric learning-based methods, some works proposed
attention-based techniques. The idea is to highlight relevant features for
detection based either on the feature itself (self-attention) or the features of
other instances (external attention). A seminal work in this field is
\cite{kang2019few}, which trains a reweighting module along with a YOLO
detector. The reweighting module outputs class-specific feature vectors through
Global Pooling (GP) from the support set. These are then channel-wise multiplied
with the query features extracted by the backbone (this operation is denoted
\textit{CRW} for Class Re-Weighting in Table \ref{tab:comparison}). Hence,
class-specific query features are generated, and the detection head computes
predictions for each class separately. Many works have been inspired by this
idea. More generally, generating class-specific features and computing the
detection separately is now standard in FSOD.

This has been used with different detection frameworks, for instance, references
\cite{li2020one} and \cite{xiao2020few} are built upon Faster R-CNN and FCOS
respectively. Other authors proposed to leverage multi-scale reweighting
vectors, such as in \cite{deng2020few}. More sophisticated ways to combine
information from query and support were also proposed. For instance, reference
\cite{fan2020fsod} trains three different heads that link, globally, locally,
and patch-to-patch the features from the query and the support. Graph Neural
Networks (GNN) can also combine the features and learn semantic relations
between classes (see e.g. \cite{xiao2020fsod, kim2020few}). Another way to
combine query and support is to simply concatenate the features as in
\cite{dai2021dynamic}, but it requires that the features have the same spatial
dimension.

\begin{figure}[h]
    \centering
    \includegraphics[width=0.9\textwidth, trim=50 0 0 0, clip]{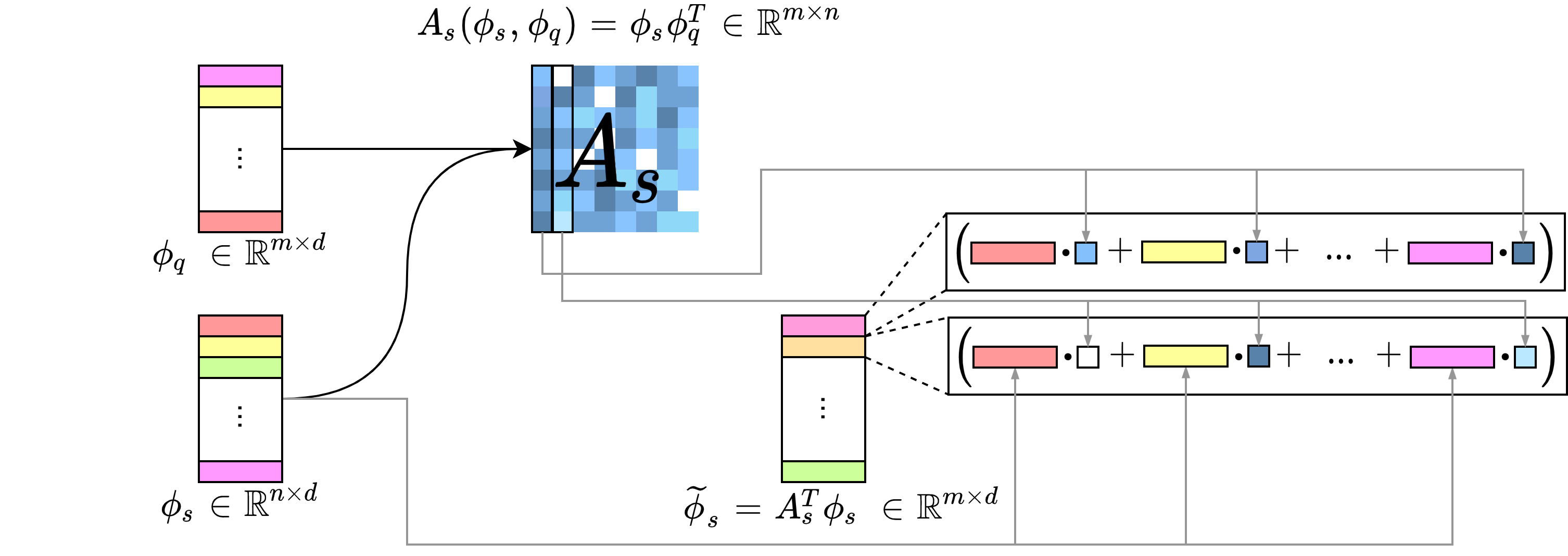}
    \caption{Spatial alignment between query and support feature maps.
     Similarity matrix is based on outer product between the maps. For sake of
     clarity, maps are reshaped as 2-D matrix where the first dimension controls
     the spatial position in the map: $m$ positions for the query and $n$ for
     the support. $d$ is the number of channels. Similar colors mean that
     features are similar.}
     \label{fig:spatial_align}
\end{figure}

Another solution proposes to adjust the size of the features maps: spatial
alignment. This consists in spatially reorganizing a feature map to match the
other. For each location in the query map, the features from one support example
(i.e. feature vectors at each location in the map) are linearly combined to
produce a feature in the aligned support map (see Figure
\ref{fig:spatial_align}). This kind of alignment has been introduced in
\cite{wang2018non} for video classification. Feature alignment between
successive frames improved overall performance by capturing long-range spatial
dependencies. This idea has already been tested for FSL as mentioned in Section
\ref{sub:basic}. Many choices can be made for the coefficients of the linear
combination, but often they are dynamically determined by computing the
similarity between query and support features (in Table \ref{tab:comparison},
all are referenced indifferently as \textit{QS Alignment} standing for
query-support alignment). References \cite{chu2021joint, chen2021should} are
based on this approach. Thus, the aligned maps can be easily combined and fed to
the detection head. The combination is often performed through a non-learnable
fusion layer composed of several point-wise operations and/or concatenation. In
Table \ref{tab:comparison}, these are denoted by their circle operator (e.g.
$\oplus$ for addition), except for concatenation and the identity
($[\cdot,\cdot]$ and \textit{Id} respectively). In some cases, additional
learnable modules are included to process the combination of query and support
features before concatenation, this is denoted as \textit{learnable} in Table
\ref{tab:comparison}. The alignment mechanism can be associated with the global
attention methods mentioned in the previous paragraph: reweighting features
globally based on the support only, as in \cite{wallach2019one} or the query and
support similarity as in \cite{han2021meta} (\textit{Global Similarity
Reweighting}). Alignment can also be carried out on the feature itself (i.e.
without information from the support) as in \cite{zhang2021meta, xu2021few}. 
The DETR framework is well suited for this kind of alignment mechanism as it is
based on Visual Transformers. The few-shot variant of DETR, Meta-DETR described
in \cite{zhang2021meta}, combines both self-attention and query-support
alignment, achieving impressive performance. It first applies self-attention
with a multi-head attention module and combines query and support features with
a single-head module where query is taken as queries and support as keys and
values.

Table \ref{tab:comparison} summarizes this literature analysis. The table is
designed to compare attention mechanisms as this is the focus of this paper.
Therefore, attention mechanisms are divided into three components that arise
from the previous review: spatial alignment, global attention, and fusion layer.
Methods without attention mechanism are also included in the Table to give a
general overview of the methods available for FSOD.
%
    \begin{table}[ht]
        \caption{Comparison of the FSOD methods from an attention perspective.
        This table separates the attention mechanisms into three components:
        spatial alignment, global attention, and fusion layer. Information about
        the original detection framework, the date of publication and the
        detection head is also included. All frameworks are working with
        multiscale features except for the one with the mention no FPN.}
        \begin{adjustbox}{width=\textwidth}
            \rowcolors{2}{gray!25}{white}
            \begin{tabular}{lllTLLT}
                \Xhline{2\arrayrulewidth}
                \textbf{Approach}                                                                    & \textbf{Name}                         & \textbf{Date}  & \textbf{Framework}         & \textbf{Alignement}                   & \textbf{Attention}          & \textbf{Fusion}                    \\ \hhline{=======}
                \cellcolor{white}                                                                    & FRW \cite{kang2019few}                & 2019           & YOLO (no FPN)              & None                                  & GP + CRW                  & None                               \\
                \cellcolor{white}                                                                    & RSI   \cite{deng2020few}              & 2019           & YOLO                       & None                                  & GP + CRW                  & None                               \\
                \cellcolor{white}                                                                    & ARMRD \cite{fan2020fsod}              & 2020           & Faster R-CNN               & None                                  & GP + CRW                  & None                               \\
                \cellcolor{white}                                                                    & VEOW  \cite{xiao2020few}              & 2020           & Faster R-CNN               & None                                  & GP + CRW                  & Pooling + $\text{Cat}\lbrack \odot, \ominus, Id \rbrack$ \\
                \cellcolor{white}                                                                    & WSAAN \cite{xiao2020fsod}             & 2020           & Faster R-CNN               & None                                  & GP + GNN + CRW            & None                               \\
                \cellcolor{white}                                                                    & CACE  \cite{wallach2019one}           & 2020           & Faster R-CNN               & QS Alignment                          & GP + CRW                  & None                               \\
                \cellcolor{white}                                                                    & KT    \cite{kim2020few}               & 2020           & Faster R-CNN               & None                                  & GP + GNN + CRW            & None                               \\
                \cellcolor{white}                                                                    & WOFT  \cite{li2020one}                & 2021           & FCOS                       & None                                  & GP + CRW                  & Pooling + Cat \textit{learnable}   \\
                \cellcolor{white}                                                                    & FPDI  \cite{gao2021fast}              & 2021           & Faster R-CNN               & Iterative Alignment via Optimization  & CRW                       & None                               \\
                \cellcolor{white}                                                                    & MFRCN \cite{han2021meta}              & 2021           & Faster R-CNN               & RoI Pooling + QS Alignment            & Global Similarity RW      & $\text{Cat}\lbrack \ominus, \text{Cat} \rbrack$ \textit{learnable}      \\
                \cellcolor{white}                                                                    & MDETR \cite{zhang2021meta}            & 2021           & DETR (no FPN)              & Transformers-based Alignment          & Transformers Self-Attention & None                               \\
                \cellcolor{white}                                                                    & DRL   \cite{liu2021dynamic}           & 2021           & Faster R-CNN               & None                                  & None                        & Pooling + $\text{Cat}\lbrack \odot, \ominus, Id \rbrack$   \\
                \cellcolor{white}                                                                    & DANA  \cite{chen2021should}           & 2021           & Faster R-CNN and RetinaNet & QS Alignment                          & Background Attenuation      & Cat                                \\
                \cellcolor{white}                                                                    & SP    \cite{xu2021few}                & 2021           & Faster R-CNN               & Self-Alignment (query and support)    & None                        & Cat                                \\
                \multirow{-15}{*}[20mm]{\cellcolor{white}\textbf{Attention}}                         & JCACR \cite{chu2021joint}             & 2021           & YOLO                       & QS Alignment         (higher order)   & None                        & Cat                                \\ \Xhline{2\arrayrulewidth}
                \cellcolor{white}                                                                    & PNPDet\cite{zhang2021pnpdet}          & 2021           & Center Net \quad(no FPN)   & None                                  & CRW                       & None                               \\
                \multirow{-2}{*}[1mm]{\cellcolor{white}\parbox{1.5cm}{\textbf{Attention/ Metric}}}   & UPE   \cite{wu2021universal}          & 2021           & Faster R-CNN               & QS Alignment                          & None                        & \textit{Id} $ \oplus$ Cat \quad \textit{learnable}                    \\ \Xhline{2\arrayrulewidth}
                \cellcolor{white}                                                                    & RM    \cite{karlinsky2019repmet}      & 2018           & Faster R-CNN               & None                                  & None                        & None                               \\
                \cellcolor{white}                                                                    & FSCE  \cite{sun2021fsce}              & 2021           & Faster R-CNN               & None                                  & None                        & None                               \\
                \multirow{-3}{*}{\cellcolor{white}\parbox{1.5cm}{\textbf{Metric Learning}}}          & PFRCN \cite{jeune2021experience}      & 2021           & Faster R-CNN               & None                                  & None                        & None                               \\ \Xhline{2\arrayrulewidth}
                \cellcolor{white}                                                                    & LSTD  \cite{chen2018lstd}             & 2018           & Faster R-CNN               & None                                  & None                        & None                               \\
                \cellcolor{white}                                                                    & WOFG  \cite{fan2021generalized}       & 2020           & Faster R-CNN               & None                                  & None                        & None                               \\
                \cellcolor{white}                                                                    & FTS   \cite{wang2020frustratingly}    & 2020           & Faster R-CNN               & None                                  & None                        & None                               \\
                \multirow{-4}{*}{\cellcolor{white}\parbox{1.5cm}{\textbf{Fine-tuning}}}              & MSPSR \cite{wu2020multi}              & 2021           & Faster R-CNN               & None                                  & None                        & None                               \\ \Xhline{2\arrayrulewidth}
                \end{tabular}%
        \end{adjustbox}
    \label{tab:comparison}
    \end{table}

\section{AAF Framework for Attention in FSOD}
\label{sec:framework}
\begin{figure}[t!]
    \centering
    \includegraphics[width=\textwidth]{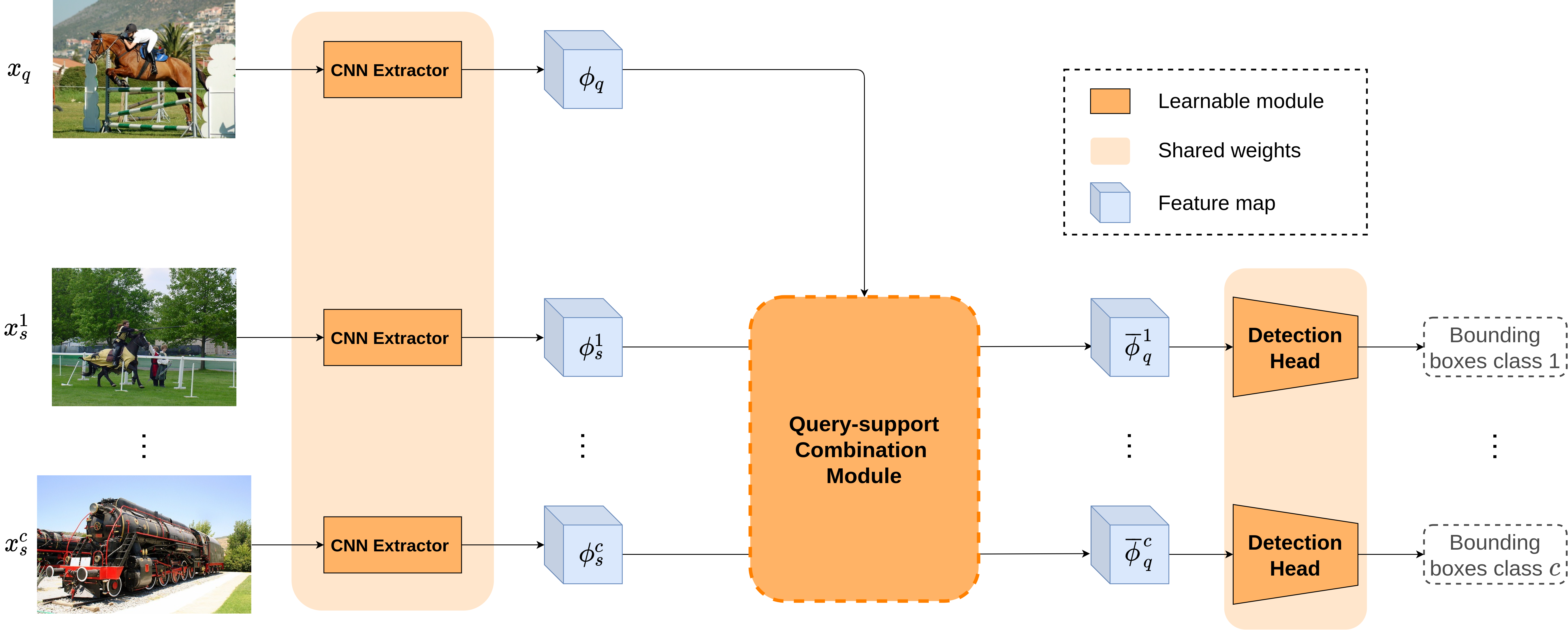}
    \caption{The main principle of attention-based FSOD method is to extract
    features from the query and support images, and combine query and support
    features to create multiple query feature maps, each specialized for the
    detection of one class. These maps are then processed by the same detection
    head to make the actual predictions.}
    \label{fig:fsod_principle}
\end{figure}

In Section \ref{sec:attention_fsod}, three main components of attention
mechanisms for FSOD have been identified. To compare them without other
interfering architecture choices, a unifying framework is proposed. The purpose
of this framework is to provide a flexible environment to represent existing
attention techniques. The three main components are: spatial alignment, global
attention and fusion layer. Most attention-based FSOD methods rely on one or
more of these components as shown in Table  \ref{tab:comparison}. Therefore, it
seems convenient to provide a flexible unifying framework that can be used to
implement these methods. The framework should take as input the features from
the query image $\phi_q$ as well as the features extracted from every support
images $\phi_s^c$ for $c \in \mathcal{C}$. It outputs class-specific query
features $\bar{\phi}_q^c$ in which features relative to class $c$ are reinforced
(see Figure \ref{fig:fsod_principle}). In order to match the three components of
attention described above, the AAF (Alignment, Attention, Fusion) framework is
also divided into three parts as shown in Figure \ref{fig:aaf}. Each component
is described below independently and examples of the possible design choices are
given. Of course, this framework is presented from the perspective of object
detection. However, it could be helpful for other kinds of few-shot tasks such
as classification, instance segmentation, or image generation. As long as a task
requires to condition the results on support examples, this framework could be
applied.

\subsection{Query Support Alignment}
The alignment module denoted  $\Lambda$, spatially aligns the features from the
query and the support. It is unlikely that objects of the same class appear at
the same position inside query and support images. Therefore, direct comparison
between their respective feature maps will not produce a high response. This is
commonly avoided by pooling the support map and using it as a class-specific
reweighting vector. This trick loses the spatial information about the support
object, which can be detrimental for the matching. Furthermore, in the case of
detection, it is likely that more than one object is visible in an image. Global
pooling is even more detrimental in this case, as it can mix features from
different objects. Instead, an alignment between query and support maps can be
performed through attention. The idea is to re-organize one feature map in
comparison to the other, so that similar features are spatially close in the
maps. The alignment operator is defined as follows:
\begin{align}
    \begin{split}
        \tilde{\phi}_q^1, ..., \tilde{\phi}_q^c, \tilde{\phi}_s^1, ..., \tilde{\phi}_s^c = \Lambda(\phi_q^1, \phi_s^1, ..., \phi_s^c)  \ \
         \ \text{with } \
          \begin{cases}
            \tilde{\phi}_q^i &= \lambda_q(\phi_q, \phi_s^i)\\
            \tilde{\phi}_s^i &= \lambda_s(\phi_s^i, \phi_q)
        \end{cases}, 
    \end{split}
\end{align}
where $\lambda_q$, $\lambda_s$ are linear combinations of their first input:
$\lambda(\phi, \rho) = A(\phi, \rho)^T\phi$ and $A \in \mathbb{R}^{m\times n}$
is an affinity matrix representing the similarity between local features from
$\phi \in \mathbb{R}^{m \times d}$ and $\rho \in \mathbb{R}^{n\times d}$. The
definition of the matrix $A$ determines how features are aligned. This
formulation is quite similar to the non-local blocks described in
\cite{wang2018non}. An example of such alignment is described in
\cite{vaswani2017attention}. Transformers attention can be understood as an
alignment of the values to match the queries-keys similarity. A major difference
is that the same feature map is used for all inputs (query, key, and value) in
visual transformers so that features are self-aligned.

As an example, Meta Faster R-CNN, described in \cite{han2021meta}, leverages an
alignment module with an affinity matrix $A(\phi, \rho) = \phi\rho^T$ which
represents the similarity between each pair of spatial locations between query
and support maps. Only the support features are aligned so that they match query
features. This can be implemented in the framework with $A_q(\phi, \rho) = I$
and $A_s(\phi, \rho) = \phi\rho^T$ (see Example A in Figure \ref{fig:aaf}).

\subsection{Global Attention}
The global attention module, denoted  $\Gamma$, combines global information of the supports
and the query. It highlights class-specific features and softens irrelevant
information for the task. This operator is defined as follows:
\begin{align}
    \begin{split}
        \hat{\phi}_q^1, ..., \hat{\phi}_q^c, \hat{\phi}_s^1, ..., \hat{\phi}_s^c = \Gamma(\tilde{\phi}_q^1, ..., \tilde{\phi}_q^c, \tilde{\phi}_s^1, ..., \tilde{\phi}_s^c) 
         \ \ \text{with} \ 
          \begin{cases}
            \hat{\phi}_q^i = \gamma_q(\tilde{\phi}_q^i, \tilde{\phi}_s^i)\\
            \hat{\phi}_s^i = \gamma_s(\tilde{\phi}_s^i, \tilde{\phi}_q^i)
        \end{cases} \mkern-18mu ,
    \end{split}
\end{align}
where $\gamma_q$, $\gamma_s$ combine the global information from their inputs
and highlight features. 
This is generally done through channel-wise multiplication.
In this way, class-specific features are highlighted, while features not
relevant to the class are softened. This module formulation is meant to be
flexible so that a wide variety of global attention mechanisms can fit into it.
For instance, reference \cite{kang2019few} pools the support maps with a global
max pooling operation ($\text{GP}$) into a reweighting vector and reweights the
query features channels with it: $\gamma(\phi, \rho) = \phi \circledast
GP(\rho)$ (see Example B in Figure \ref{fig:aaf}).

\subsection{Fusion Layer}
The purpose of the fusion component is to combine query and support maps. This
is only applicable when the maps have the same spatial dimension. It is mostly
used alongside with the alignment module, which does not combine the information
from the support and the query but only reorganize the maps according to the
other. In particular, when support and query maps do not have the same spatial
dimension, aligning support to query can fix the size discrepancy. The fusion
operator is defined as follows:
\begin{align}
    \begin{split}
        \bar{\phi}_q^1, ..., \bar{\phi}_q^c = \Omega(\hat{\phi}_q^1, ..., \hat{\phi}_q^c, \hat{\phi}_s^1, ..., \hat{\phi}_s^c) \ \
         \ \text{with } \
         \bar{\phi}_q^i = \omega(\hat{\phi}_q^i, \hat{\phi}_s^i),
    \end{split}
\end{align}
where $\omega$ is generally the concatenation of the results of multiple point-wise
operators: $\omega(\phi, \rho) = \omega_1(\omega_2(\phi, \rho), ...,\omega_r(\phi,
\rho))$ with $\omega_i \in \{\oplus, \odot, \ominus, [\cdot,\cdot], ...\}$.

\begin{figure}[ht]
    \centering
    \includegraphics[width=\textwidth]{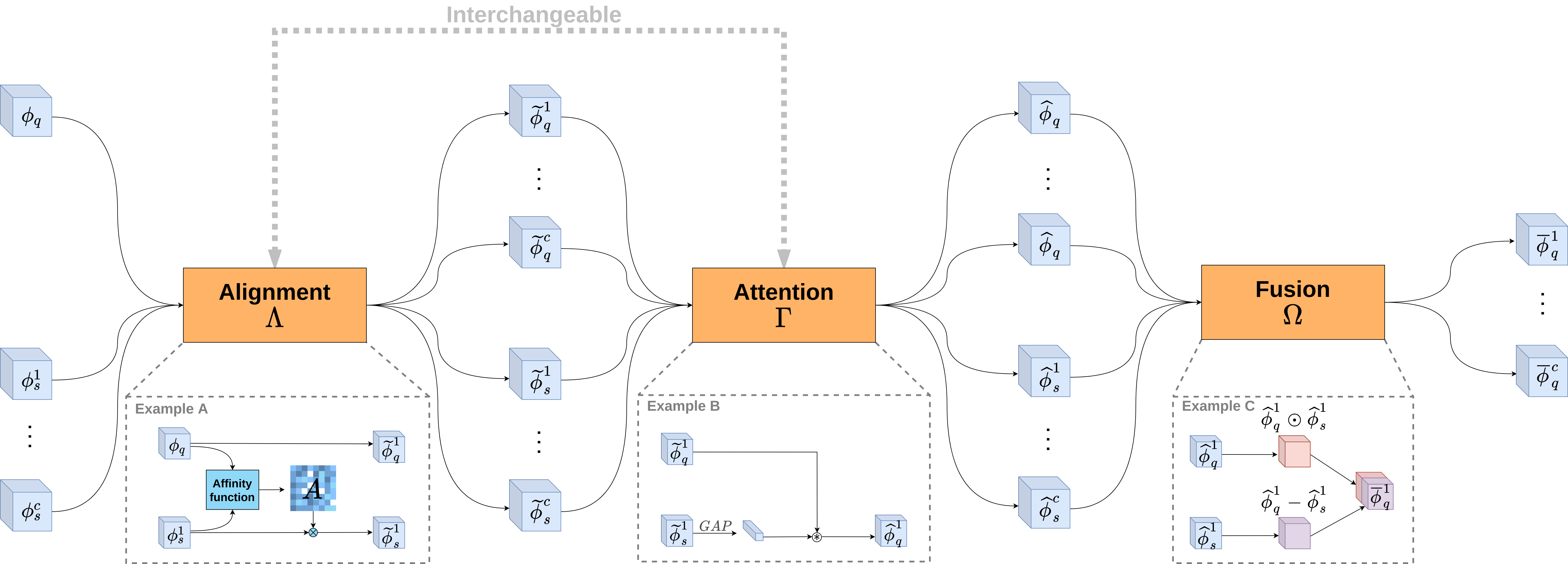}
    \caption{The Alignment Attention Fusion (AAF) module is composed of
    three components: spatial alignment $\Lambda$, global attention $\Gamma$
    and a fusion layer $\Omega$. Examples for each module are depicted,
    these come from FSOD methods in the literature. Example A is presented
    in \cite{han2021meta}, Example B in \cite{kang2019few} and Example C in \cite{liu2021dynamic}.}
    \label{fig:aaf}
\end{figure}

An example of such fusion module is presented in \cite{liu2021dynamic} in which
the query and support feature maps are element-wise multiplied, subtracted and
then concatenated together: $\omega(\phi, \rho) = [\phi \odot \rho, \phi \ominus
\rho]$ (see Example C in Figure \ref{fig:aaf}). The point-wise operators
can also contain small trainable models such as in \cite{han2021meta}, where small
CNNs are applied after the point-wise operator, but before the concatenation:
$\omega(\phi, \rho) = [\psi_{dot}(\phi \odot \rho), \psi_{sub}(\phi
\ominus \rho), \psi_{cat}([\phi, \rho])]$. \\

The overall framework results in the successive application of the three
components presented above:

\vspace{-4mm}
\begin{equation}
    \bar{\phi}_s^1, ..., \bar{\phi}_s^c = \text{AAF}(\phi_q^1, \phi_s^1, ..., \phi_s^c) = (\Omega \circ \Gamma \circ \Lambda)(\phi_q^1, \phi_s^1, ..., \phi_s^c).
\end{equation}

\noindent
Except for the fusion layer which must be applied last, spatial alignment and
global attention can be applied in any order. In most cases, it will produce
different results as they are probably not commutative. However, some FSOD
methods (e.g. \cite{chen2021should}) require applying global attention before
spatial alignment, so the framework must be flexible enough to model this as
well. The whole architecture of the AAF framework is illustrated in Figure
\ref{fig:aaf}, in which examples from the previous sections are also depicted.

\section{Attention-based FSOD Comparison}
\label{sec:comparison}
In order to showcase the flexibility of the proposed AAF module, a comparison
between multiple existing works is conducted. Some methods described in Section
\ref{sec:related} are selected: FRW \cite{kang2019few}, WSAAN
\cite{xiao2020fsod}, DANA \cite{chen2021should}, MFRCN \cite{han2021meta} and
DRL \cite{liu2021dynamic} (see Table \ref{tab:comparison}). These have been
chosen because they represent well the variety of attention mechanisms available
in the literature. FRW is based on class-specific reweighting vectors, WSAAN has
a more sophisticated global attention and computes reweighting vectors inside a
graph structure. DANA and MFRCN leverage query-support alignment in a slightly
different manner and DRL only uses a sophisticated fusion layer. Each of these
methods has been reimplemented within the AAF framework. Of course, some details
differ from the original implementations, but the purpose of this comparison is
to evaluate only the attention mechanisms. In particular, the backbone and the
training strategy (losses and episode tasks) may differ. The main comparison is
done on Pascal VOC dataset \cite{everingham2010pascal} following the literature
on FSOD. Some results are also provided on COCO dataset \cite{lin2014microsoft}.

\subsection{AAF Framework Implementation Details}
In order to make the comparison fair, some implementation details are kept fixed
for all experiments. The backbone is a ResNet-50 with a 4-layers Feature Pyramid
Network (FPN) on top \cite{lin2017feature}. It extracts features at 4 different
levels, which should help the network to detect objects at multiple scales. As
features are extracted at multiple levels, attention mechanisms are also
implemented to work at different scales. This may differ from the original
implementation, but most methods are designed to work at multiscale (see Table
\ref{tab:comparison}). The networks are trained in an episodic manner. During
each episode, a subset $\mathcal{C}_{ep} \subset \mathcal{C}$ of the classes is
randomly sampled ($\lvert\mathcal{C}_{ep} \rvert= 5$). Only the annotations of
these classes are shown to the networks. It means that the networks are not
penalized for not detecting other classes. An episode is constituted of 100
images for each class, which constitutes a query set of 500 images. The base
training is made of 1000 (3000 for COCO) of such episodes. With a batch size of
8, this corresponds to 60k (180k for COCO) iterations. The optimization is
performed with SGD and a learning rate of $\num{1e-3}$. The support set is
composed of one example per class (i.e. one image with only one annotation) and
a new support set is sampled at each iteration. Next, a fine-tuning stage occurs
with a smaller learning rate ($\num{1e-4}$). During this phase, test classes are
added to the class selection. For these classes only $k$ examples are available
($k \in \{1,3,5,10\}$). Hence, the query and support sets are both composed of
$5k$ images. Images for base classes change between episodes while for test
classes, they remained fixed. To keep the training comparable between multiple
values of $k$, the number of episodes is adjusted so that the same number of
weight updates is performed with each configuration. The selected methods are
compared on Pascal VOC dataset, following the setup detailed in
\cite{kang2019few}. Training images are selected from VOC 2007 and 2012 train
and validation set while the evaluation is performed on VOC 2007 test set. Five
classes among the 20 available have been selected as test classes and have only
been seen by the networks during the fine-tuning. For experiments on COCO, 20
classes are selected as test classes. As in the FSOD literature, these classes
are the ones from Pascal VOC dataset. These details are often different from one
method to another for FSOD. This makes the comparison cumbersome between
different works. The purpose of this framework is to facilitate the
implementation of new attention techniques while providing a fair way to compare
them.

\subsection{Results and Analysis on Pascal VOC}
\label{sec:res_voc}
In Table \ref{tab:result_voc}, mean Average Precision (mAP) is reported for each
of the selected methods from Section \ref{sec:comparison}. mAP is computed with
a $0.5$ overlap threshold. For each method, the evaluation is split into base
and novel classes. Results on base classes demonstrate the ability of the method
to solve object detection with a large dataset available. This should be
compared to the performance of vanilla FCOS on the same classes ($0.68$). While
it is important to keep performance on base classes close to the baseline, the
most important metric for FSOD is performance on novel classes. Evaluation
on novel classes shows how well a method generalizes to unseen classes with
limited data.

As mentioned in \cite{huang2021survey}, two separate ways to evaluate the
performance on novel classes exist in the FSOD literature. The first one,
introduced by \cite{kang2019few} samples only one fixed support set at test time
(the same that is used during fine-tuning). The second one consists in repeating
this process with multiple different support. The former is less reliable and
often overestimates the generalization capabilities. The latter is a lot more
time-consuming as it requires at least 30 different runs (i.e. fine-tuning and
evaluation) to become steady. Results from Table \ref{tab:result_voc} are
computed using the quicker but less reliable method. However, the goal of this
work is to provide a tool that helps to conduct a fair comparison between
different FSOD techniques. Therefore, it must also adopt fair and reliable
metrics. It is planned to perform more reliable comparisons as future work (i.e.
repeating with different support sets and class splits).

\begin{table}[ht]
    \caption{Performance comparison between five selected methods (see Section
    \ref{sec:comparison}). All are reimplemented with the proposed AAF
    framework. Mean average precision is reported for each method on base and
    novel classes separately and for various numbers of shots (1, 3, 5 and 10).
    Bold values indicate the best performing method for each number of shots and
    for base/novel classes separately.}
    \label{tab:result_voc}
    \resizebox{\textwidth}{!}{%
    \begin{tabular}{c|ccccccrrcc}
    \hline
            & \multicolumn{2}{c}{\textbf{FRW} \cite{kang2019few}} & \multicolumn{2}{c}{\textbf{WSAAN} \cite{xiao2020fsod}} & \multicolumn{2}{c}{\textbf{DANA} \cite{chen2021should}} & \multicolumn{2}{c}{\textbf{MFRCN} \cite{han2021meta}}  & \multicolumn{2}{c}{\textbf{DRL} \cite{liu2021dynamic}} \\ \hline
    \# Shots & \underline{Base}            & \underline{Novel}          & \underline{Base}        & \underline{Novel}                & \underline{Base}            & \underline{Novel}           & \underline{Base}                     & \underline{Novel}    & \underline{Base}                & \underline{Novel}      \\
    1       & 0.599           & 0.282          & 0.617       & 0.309                & 0.626           & \textbf{0.328}  & 0.578                    & 0.302    & \textbf{0.642}      & 0.270      \\
    3       & 0.633           & 0.311          & 0.635       & \textbf{0.422}       & \textbf{0.642}  & 0.340           & 0.587                    & 0.368    & 0.617               & 0.296      \\
    5       & 0.643           & 0.463          & 0.647       & \textbf{0.462}       & 0.652           & 0.426           & 0.621                    & 0.408    & \textbf{0.664}      & 0.373      \\
    10      & 0.632           & 0.487          & 0.653       & \textbf{0.517}       & 0.650           & 0.503           & 0.634                    & 0.494    & \textbf{0.670}      & 0.480      \\ \hline
    \end{tabular}%
    }
    \vspace{-1mm}
    \end{table}

\begin{figure}[ht]
    \centering
    \includegraphics[width=0.95\textwidth, trim=0 30 0 20,clip]{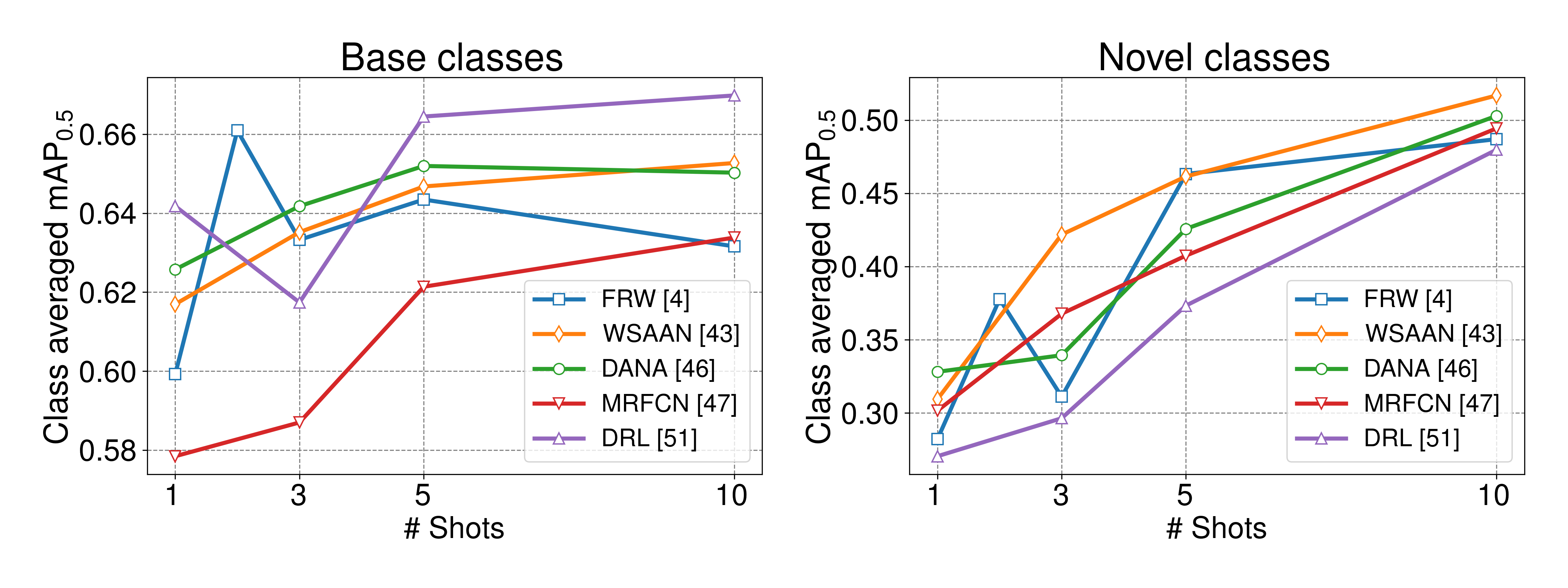}
    \caption{Evolution of $\text{mAP}_{0.5}$ with the number of shots averaged
            on base and novel classes. Each line represents one of the
            reimplemented methods. Increasing the number of examples improve
            performance of the model. This improvement is larger for novel
            classes as the network has not been trained (apart from fine-tuning)
            to detect them.}
    \label{fig:method_compare}
    \vspace{-2mm}
\end{figure}

\begin{figure}[t]
    \centering
    \includegraphics[width=\textwidth, trim=0 0 0 5,clip]{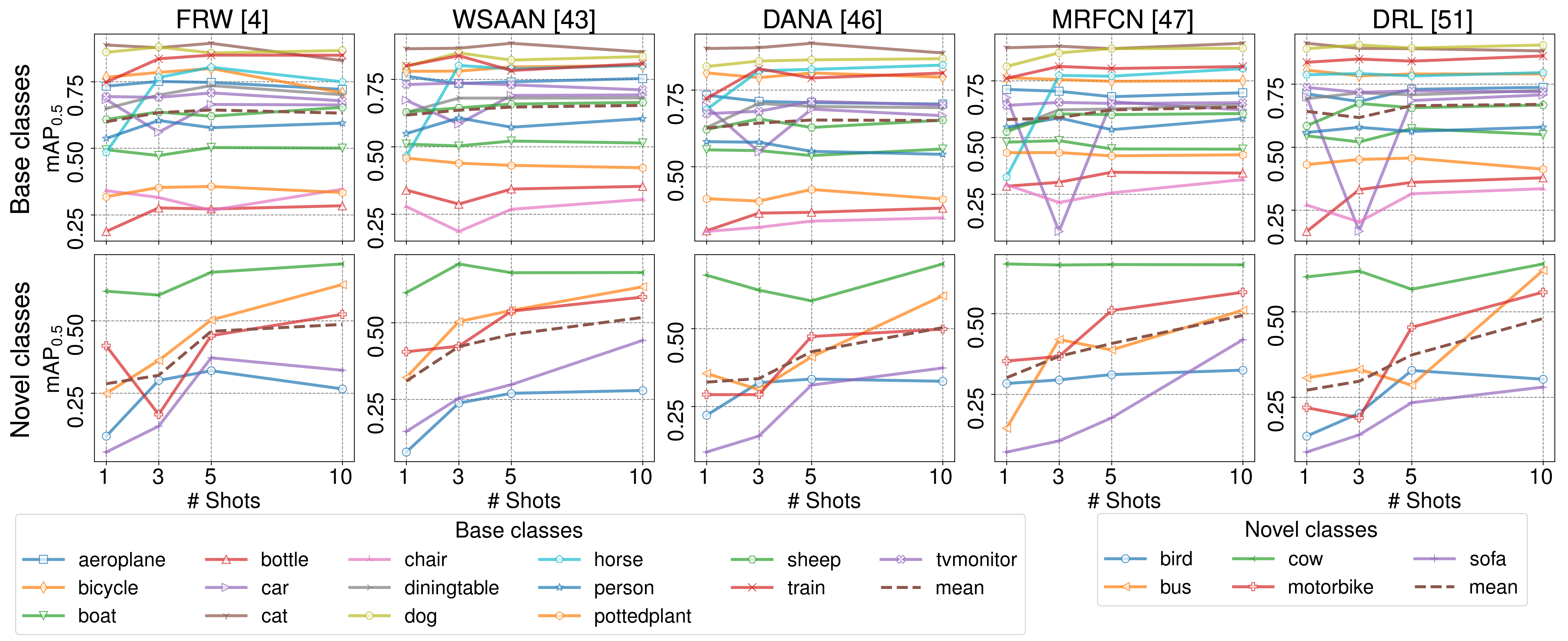}
    \caption{$\text{mAP}_{0.5}$ on Pascal VOC against the number of shots for
            each class and each method. Dashed lines represent average
            performance on all classes, either base classes(top row) or novel
            classes(bottom row).}
    \label{fig:map_per_class}
\end{figure}

From Table \ref{tab:result_voc}, one can observe, on base classes, a slight
decrease in performance compared to FCOS baseline. This is expected, even if the
network has seen a lot of examples of these classes during training, its
predictions are still conditioned on a few examples, which can sometimes be
misleading. On the other hand, performance on novel classes is significantly
lower than the FCOS baseline (i.e. trained with plenty of examples), especially
for low numbers of shots. The number of shots is crucial for performance on
novel classes. The higher the number of shots, the better the network performs.
In average, with 10 examples per class, the network achieves $0.2$ higher mAP
than with 1 example. More examples provide more precise and robust class
representations, improving detection. The same phenomenon is observed with base
classes with a smaller magnitude ($+ 0.04$ mAP from 1 to 10 shots). Figure
\ref{fig:method_compare} displays these trends clearly, both for base and novel
classes. More detailed results are given in Figure \ref{fig:map_per_class} where
the evolution of performance per class is plotted. From that, one can notice
that performance varies greatly from a class to another. Classes that are the
most difficult benefit the most from having more examples. This is true both for
base and novel classes. In addition, the novel classes that are easier to detect
are often quite similar to some base classes (e.g. \textit{cow} is similar to
\textit{horse} and \textit{sheep}), thus requiring fewer examples to achieve
good detection.

This behavior is expected from any few-shot object detection method. Moreover,
performance values are close to what is reported in the original papers. Of
course, these are not the exact same values as many architectural choices differ
from the proposed methods (e.g. backbone, classes splits, losses, etc.).
Nevertheless, it confirms that the proposed AAF framework is flexible enough to
implement different attention mechanisms. Therefore, it is an appropriate tool
to compare different methods and to find what are the most efficient operations
to combine query and support information for FSOD.

DRL is arguably the simplest method among the five selected as it leverages
only a fusion layer. It combines query features with the features of each
support image through concatenation and point-wise operations, creating
class-specific query features. It is therefore the closest to the regular FCOS
functioning. This explains the very good performance (compared to the baseline)
on base classes and lower mAP on novel classes. Regarding the other methods, FRW
and WSAAN can be easily compared as both are based on global attention. The only
difference is how the class-specific vectors are computed. In FRW, they are
globally pooled from the support feature map. However, WSAAN combines the same
vector with query features in a GNN. This certainly provides more adaptive
class-specific features yielding better results both on base and novel sets.
The remaining methods, DANA and MFRCN both leverage spatial alignment. While it
seems to bring quite an improvement for DANA over FRW or DRL, the gain is
smaller for MFRCN. In both methods, spatial alignment is not used alone. It is
combined with other attention mechanisms. In DANA, a Background Attenuation
block (i.e. a global self-attention) is applied to the support features to
highlight class-relevant features and soften background ones. In MFRCN, aligned
features are reweighted with global vectors computed from the similarity matrix
between query and support features. This last operation may be redundant as the
similarity information is already embedded into the aligned features, whereas
background attenuation could extract new information. To confirm this
hypothesis, a specific ablation study on both of these methods is planned as
future work.

From this comparison, one can conclude that both global attention and spatial
alignment are beneficial for FSOD. However, these improvements may not always be
compatible. Hence, the design of each component must be done carefully so that
spatial alignment, global attention, and fusion work in unison.

\subsection{Results and Analysis on COCO}

Another set of experiments is conducted on COCO dataset. Only the two
best-performing methods from Section \ref{sec:res_voc} are selected and trained
on COCO following the same experimental setup. The results are summarized in
Table \ref{tab:result_coco}. The mAP value are reported following standards from
Pascal VOC ($\text{mAP}_{0.5}$ with one IoU threshold), and COCO
($\text{mAP}_{0.5:0.95}$ with several thresholds). This is done for 1, 5, 10,
and 30 shots. These results comfort the conclusion obtained on Pascal VOC: the
framework is flexible enough to implement various FSOD techniques that achieve
competitive results with state-of-the-art. As on Pascal VOC the networks achieve
better detection with more shots. While more beneficial for novel classes, base
classes also benefit significantly from a higher number of examples, unlike on
Pascal VOC. WSAAN outperforms DANA on Pascal VOC but performs slightly worse on
COCO.

\begin{table}[]
    \centering
    \caption{Performance comparison between the two best performing methods from
                Section \ref{sec:res_voc} on COCO. $\text{mAP}_{0.5:0.95}$ (COCO
                mAP, with IoU thresholds ranging from 0.5 to 0.95) and
                $\text{mAP}_{0.5}$ values are reported in the table for base and
                novel classes separately and for different numbers of shots.}
    \label{tab:result_coco}
    \resizebox{0.9\textwidth}{!}{%
    \begin{tabular}{c|cccc|cccc}
    \hline
           & \multicolumn{4}{c|}{\textbf{WSAAN} \cite{xiao2020fsod}}                                                     & \multicolumn{4}{c}{\textbf{DANA} \cite{chen2021should}}                                                           \\
           & \multicolumn{2}{c}{$\text{mAP}_{0.5}$}                      & \multicolumn{2}{c|}{$\text{mAP}_{0.5:0.95}$}  & \multicolumn{2}{c}{$\text{mAP}_{0.5}$}                       & \multicolumn{2}{c}{$\text{mAP}_{0.5:0.95}$}        \\ \hline
    \# Shot& \underline{Base}   & \multicolumn{1}{c|}{\underline{Novel}} & \underline{Base}    & \underline{Novel}       & \underline{Base}   & \multicolumn{1}{c|}{\underline{Novel}}  & \underline{Base}       & \underline{Novel}         \\
    1      & 0.335              & \multicolumn{1}{c|}{0.120}             & 0.201               & 0.066                   & \textbf{0.355}     & \multicolumn{1}{c|}{\textbf{0.145}}     & \textbf{0.213}         & \textbf{0.078}            \\
    5      & 0.399              & \multicolumn{1}{c|}{0.199}             & 0.236               & 0.105                   & \textbf{0.428}     & \multicolumn{1}{c|}{\textbf{0.222}}     & \textbf{0.252}         & \textbf{0.119}            \\
    10     & 0.409              & \multicolumn{1}{c|}{0.214}             & 0.244               & 0.115                   & \textbf{0.430}     & \multicolumn{1}{c|}{\textbf{0.237}}     & \textbf{0.256}         & \textbf{0.129}            \\
    30     & 0.415              & \multicolumn{1}{c|}{0.222}             & 0.247               & 0.121                   & \textbf{0.435}     & \multicolumn{1}{c|}{\textbf{0.244}}     & \textbf{0.260}         & \textbf{0.133}            \\ \hline
    \end{tabular}%
    }
    \end{table}

\section{Conclusion and Future Work}
\label{sec:conclusion}
This paper introduces a modular framework that facilitates the implementation of
different attention mechanisms for FSOD. Thanks to this, a fair comparison
of the various techniques has been conducted. The experiments carried out in
this work demonstrate the flexibility of the proposed framework and prove that
it is convenient for comparing attention techniques. More comparisons are
planned as future work, along with in-depth ablation studies and the design of
new attention techniques for FSOD. To help the development of such methods and
future comparisons, the code of the proposed AAF framework will be made
available.

\bmhead{Acknowledgments}
The authors would like to thank COSE for their close collaboration and the
funding of this project.

\singlespacing
\bibliography{bibliography}



\end{document}